\pdfoutput=1

\documentclass[11pt]{article}

\usepackage[]{ACL2023}

\usepackage{times}
\usepackage{latexsym}
\usepackage{amsmath}
\usepackage{amssymb}
\usepackage{graphicx}
\usepackage{algorithm}
\usepackage{algpseudocode}
\usepackage{booktabs}
\usepackage[T1]{fontenc}

\usepackage[utf8]{inputenc}

\usepackage{microtype}

\usepackage{inconsolata}

\title{Mimicking How Humans Interpret Out-of-Context Sentences Through Controlled Toxicity Decoding}

\author{Maria Mihaela Trusca$^*$ \and Liesbeth Allein$^*$ \\
         Department of Computer Science \\ KU Leuven \\ \texttt{firstname(s).lastname@kuleuven.be}}

\begin{document}
\maketitle
\def\thefootnote{*}\footnotetext{Equal contribution.}\def\thefootnote{\arabic{footnote}}
\begin{abstract}
Interpretations of a single sentence can vary, particularly when its context is lost. This paper aims to simulate how readers perceive content with varying toxicity levels by generating diverse interpretations of out-of-context sentences. By modeling toxicity, we can anticipate misunderstandings and reveal hidden toxic meanings. Our proposed decoding strategy explicitly controls toxicity in the set of generated interpretations by (i) aligning interpretation toxicity with the input, (ii) relaxing toxicity constraints for more toxic input sentences, and (iii) promoting diversity in toxicity levels within the set of generated interpretations. Experimental results show that our method improves alignment with human-written interpretations in both syntax and semantics while reducing model prediction uncertainty.
\end{abstract}

\section{Introduction}

Misunderstandings online can often be traced back to misalignment between the meanings of text intended by the author and those inferred by the readers. This is even further amplified when text is taken out of context -- which is commonplace on social media -- resulting in frustration and heated discussion. In this paper, we aim to mimic how readers may interpret out-of-context sentences. We do this by modeling and generating for each sentence a diverse set of interpretations \citep{DBLP:journals/ai/AlleinTM25}. Toxicity is taken as the control factor during generation as we want to simulate human interpretation behavior of sentences that exhibit varying degrees of surface-level toxicity.
Generating diverse interpretations can help anticipate misunderstandings, explain reactions from readers, and recover underlying toxicity, which is especially beneficial for capturing implicit hostility and harm online \citep{elsherief-etal-2021-latent,hartvigsen-etal-2022-toxigen}. 

This paper introduces \textit{a novel decoding strategy for interpretation generation that explicitly controls the toxicity level of generated interpretations.} Our decoding strategy enforces three key objectives that are inspired by toxicity patterns observed in human-written sentence interpretations: \textit{Align the toxicity level} of generated interpretations with that of the input sentence (Objective 1); \textit{Progressively relax toxicity constraints} on the interpretations for increasing toxicity in the sentence (Objective 2); \textit{Promote diversity in the toxicity levels} across the generated interpretations (Objective 3). These objectives are enforced iteratively during the decoding process, enabling fine-grained control over toxicity while maintaining coherence and diversity in generated text. Controlling generation in the decoding phase is particularly desirable as it bypasses the need for alterations to model architectures, allowing a plug-and-play integration with existing language models.


Our results demonstrate the soundness and effectiveness of our decoding strategy. Controlling the decoding of interpretations using all three objectives consistently leads to generated interpretations that better align with human-written interpretations in terms of syntax and semantics, compared to when generation is not controlled. Our strategy also lowers uncertainty for the base models when predicting the interpretations. 

\section{Related Work}

Text generation can be controlled using a range of control factors, including text attributes (e.g., sentiment, style) \citep{hu2017toward,dathathriplug}, syntactic structures \cite{li2022diffusion}, speaker or reader characteristics \cite{dinan-etal-2020-queens,majumder-etal-2020-like}, and structured data (e.g., tables, knowledge graphs) \cite{zhang2023survey}. A popular approach to condition text generation is in-context learning, where these control factors are integrated into the input \cite{yang-etal-2023-tailor}. Another method is to control generation during the decoding phase, e.g., by manipulating the output token distributions \cite{pascual-etal-2021-plug-play, yang-klein-2021-fudge, kim-etal-2023-critic}. 

This paper controls the toxicity of generated interpretations based on the surface-level toxicity of the original sentence during decoding. While much of the existing work on controlling toxicity in text generation focuses on reducing toxicity \cite{gehman-etal-2020-realtoxicityprompts, liu-etal-2021-dexperts,prabhumoye-etal-2023-adding,wingate-etal-2022-prompt}, our work builds on the idea that the toxicity of the original sentence is perceived differently among readers. We aim to capture this variability by constraining generation following three objectives.


\section{Methodology}


\subsection{Preliminaries}\label{sec:preliminaries}

Language models generate text sequences $y$ of length $T$ by decoding the probability of the sequence $y$ calculated using the chain rule: $p(y) = \prod_{t=1}^{T}{p(y_t|y_{<t})}$, where $y_{<t}=\{y_1, ..., y_{t-1}\}$. The probabilities $p(y_t|y_{<t})$ are obtained by projecting the logits computed by the language model into the space of the model's vocabulary $\mathcal{V}$ typically using a softmax transformation. By applying the logarithmic differentiation over the chain rule, the \textit{softmax} scores are given by $score(y_t|y_{<t})=\log{p(y_t|y_{<t})}$. Once the scores are computed, a decoding algorithm such as nucleus sampling or beam search is applied to autoregressively generate $y$. 

In our work, we aim to control the toxicity of the interpretations generated by a language model for an input sentence in a plug-and-play manner. We do this by calibrating the \textit{softmax} scores for toxicity control before applying the decoding algorithm. To ensure the correct summation of all probabilities in the $\mathcal{V}$ space to 1, we apply another \textit{softmax} transformation over the calibrated scores.

\subsection{Toxicity control}\label{sec:toxicity}

We define a set of objectives for our decoding strategy that should closely align the generated interpretations with the toxicity behavior observed in the input sentence and human interpretations. The implementation of these objectives is summarized in Algorithm \ref{alg:alg_1}.

\begin{algorithm}
\caption{The implementation of Objectives 1-3}\label{alg:alg_1}
\footnotesize
\textbf{Input} {$s$, $tox(s)$, $tox(y_t) \in \mathbb{R}^V$, $y=\{\}$}\\
\textbf{Output} {$y$}
\begin{algorithmic}
\If{Objective 3 and $(\exists) y^{'}$} 
     \If{$tox(y^{'}) < tox(s)$}
     \State $tox(s) = tox(s) + (tox(s)-tox(y^{'}))$
     \ElsIf{$tox(y^{'}_{T}) > tox(s)$}{}
     \State $tox(s) = tox(s) - (tox(y^{'}) - tox(s))$
     \EndIf
 \EndIf
 \While{$t \leq T$} 
 \State Compute $score(y_t|y_{<t})$
 \If{Objective 1}
     \If{Objective 2}
     \State $\lambda=1/(tox(s)*100)$
     \ElsIf{not Objective 2}{$\lambda=1$}
     \EndIf
     \If{$tox(y_{<t}) < tox(s)$} 
     \State$ score(y_t|y_{<t}) = score(y_t|y_{<t}) +$
     \State $\lambda * tox(y_t)$
     \EndIf
     \If{$tox(y_{<t}) > tox(s)$} 
     \State$ score(y_t|y_{<t}) = score(y_t|y_{<t}) -$
     \State $\lambda * tox(y_t)$
     \EndIf
 \EndIf
 \State $y^{*}_t=argmax(score(y_t|y_{<t}))$
 \State $y_{<t} = y_{<t} + y^{*}_t$
 \EndWhile
 \If{t=T}
 \State $y=y_{<t}$
 \EndIf
 \end{algorithmic}
 \end{algorithm}

\paragraph{Objective 1: Match toxicity level of the interpretations to the sentence} 
The toxicity of the generated interpretations should match the toxicity level of the input sentence, as maintaining consistency in toxicity prevents the interpretations from unintentionally intensifying or minimizing the original tone. Adopting this hypothesis, we ensure that the generated interpretation preserves the meaning of the input sentence in terms of toxicity.
Since the text generation process is sequential, it is necessary to calibrate the toxicity level of the generated text after each time step $t$.

Knowing that the $tox(*)$ function indicates the toxicity level (the codomain of the function is $[0,1]$) and given the \textit{softmax} scores $score(y_t \mid y_{<t})$ computed by the language model for the $t$-th generated token $y_t$ based on the already generated sequence of $t-1$ tokens $y_{<t}$, we calibrate the scores as follows:
\begin{multline}\label{eq:eq_1}
  score(y_t|y_{<t}) = score(y_t|y_{<t}) + \lambda * tox(y_t) \text{,}\\
  \hspace{3cm} \text{if } tox(y_{<t}) < tox(s)\\
  score(y_t|y_{<t}) = score(y_t|y_{<t}) - \lambda * tox(y_t) \text{,}\\
  \text{if } tox(y_{<t}) > tox(s)
\end{multline}
where $s$ is the input sentence, $\lambda$ adjusts the toxicity control, and $tox(y_t)\in$ $\mathbb{R}^\mathcal{V}$ indicates the toxicity level of $y_t$ in $\mathcal{V}$ used by the language model. All toxicity scores are computed using the well-established BERT-HateXplain model \cite{DBLP:conf/aaai/MathewSYBG021}. 

By implementing \textbf{Objective 1} using Eq. \ref{eq:eq_1}, we correct the toxicity of the generated interpretation after every time step $t$ to ensure that the toxicity of the final interpretation converges to that of the input sentence.

\begin{table}[t!]
    \centering
    \scriptsize
    \begin{tabular}{cc}
    \toprule
       \textbf{Toxicity Interval} &  \textbf{Toxicity Average Standard}  \\
       \textbf{of the Input Sentence} & \textbf{Deviation of the Interpretations} \\
      \midrule
    (0.0 - 0.2) & 0.05 \\
    (0.2 - 0.4) & 0.10 \\
    (0.4 - 0.6) & 0.13 \\
    (0.6 - 0.8) & 0.20 \\
    (0.8 - 1.0) & 0.23 \\

    \bottomrule
    \end{tabular}
    \caption{Comparison between the toxicity intervals of the input sentences and the average standard deviations of the toxicity scores of all interpretations per input sentences. The average is computed at the interval level. 
    }\label{tab:tab_0}
\end{table}


\paragraph{Objective 2: Gradually relax control as sentence toxicity rises} Empirically, we observe that input sentences with higher toxicity scores are more likely to have human interpretations with a broader toxicity range than less toxic input sentences. As shown in Table \ref{tab:tab_0}, the standard deviation of the toxicity scores observed in the human interpretations of an input sentence is higher for more toxic input sentences than for less toxic ones. 
Based on this observation, we gradually loosen the toxicity control over the generated interpretations as the toxicity of the input sentence increases. To implement this, we set the weight $\lambda$ in Eq. \ref{eq:eq_1} as $1 / ({tox}(s) \cdot 100)$. If \textbf{Objective 2} is not implemented, $\lambda$ is set to 1.

\paragraph{Objective 3: Promote diversity by alternating toxicity} While the generated interpretations should preserve the meaning of the input sentence, we also want to capture the range of possible interpretations. To encourage diversity in the set of generated interpretations, we set a heuristic rule that the current generated interpretation should be higher in toxicity than the input sentence when the previous interpretation was lower in toxicity, and vice versa. To implement this, we update the toxicity score of the input sentence, $tox(s)$, after every generated interpretation as follows:
\begin{multline}\label{eq:eq_2}
    \hspace{0.5cm} tox(s) = tox(s) + (tox(s)-tox(y^{'}))\text{,} \\
   \hspace{3cm} \text{if } tox(y^{'}) < tox(s) \\
    tox(s) = tox(s) - (tox(y^{'}) - tox(s))\text{,} \\
    \text{if } tox(y^{'}) > tox(s) 
\end{multline}
where $y^{'}$ is the previously generated interpretation.

Note that our decoding strategy defines the toxicity of interpretations as a function of the input sentence toxicity, meaning that we can always substitute the toxicity score of the input sentence with an arbitrary value. This feature is particularly important for content moderation by producing interpretations that deliver the meaning of the input sentence in a non-toxic manner.

\section{Experimental Setup}\label{sec:setup}

\paragraph{Dataset} We rely on the OrigamIM dataset\footnote{\url{https://github.com/laallein/origamIM}.} \cite{allein-moens-2024-origamim} to evaluate our decoding strategy. OrigamIM is the first dataset that specifically supports the interpretation modeling task \cite{DBLP:journals/ai/AlleinTM25} and includes 9,851 human-written interpretations of 2,018 sentences from Reddit posts. To accommodate the language models for this task, we fine-tune and validate them on the OrigamIM training and validation sets. The test set is used to evaluate our decoding strategy.

\paragraph{Models} To evaluate our method for toxicity control, we use three open-source language models: BART (139M parameters) \cite{DBLP:conf/acl/LewisLGGMLSZ20}, T5 (223M parameters) \cite{DBLP:journals/jmlr/RaffelSRLNMZLL20}, and LLAMA 7b (6.74B parameters) \cite{DBLP:journals/corr/abs-2302-13971}. We test various combinations of our proposed objectives and compare it against the base models without explicit control. 


\begin{table*}[t!]
    \centering
    \footnotesize
    \begin{tabular}{l cccc}
    \toprule
       Method & \multicolumn{1}{c}{$\text{$METEOR$}(\uparrow$)}  &  \multicolumn{1}{c}{$\text{$COMET$}(\uparrow$) } & \multicolumn{1}{c}{$\text{$Perplexity$}(\downarrow$)}  &  \multicolumn{1}{c}{$\text{$Correlation$}(\uparrow$) }  \\
      \midrule
$BART$ & 29.22 $\pm$ 0.21 & 82.36 $\pm$ 0.31 & 1.27 $\pm$ 0.2 & 0.43 $\pm$ 0.56 \\
$BART+Obj_{1}$ & \textbf{29.82 $\pm$ 0.12} & 83.74 $\pm$ 0.21 & 1.27 $\pm$ 0.1 & 0.41 $\pm$ 0.49 \\
$BART+Obj_{1,2}$ & 29.48 $\pm$ 0.23 & 83.11 $\pm$ 0.3 & \textbf{1.26 $\pm$ 0.1} & 0.45 $\pm$ 0.23 \\
$BART+Obj_{1,3}$ & 29.01 $\pm$ 0.22 & 84.16 $\pm$ 0.36 & \textbf{1.26 $\pm$ 0.2} & 0.42 $\pm$ 0.31 \\
$BART+Obj_{1,2,3}$ & 29.79 $\pm$ 0.12 & \textbf{85.81 $\pm$ 0.37} & 1.27 $\pm$ 0.1 & \textbf{0.46 $\pm$ 0.34} \\
\midrule
$LLAMA$ & 27.13 $\pm$ 0.44 & 86.16 $\pm$ 0.26 & 13.19 $\pm$ 0.3 & 0.41 $\pm$ 0.32 \\
$LLAMA+Obj_{1}$ & 27.73 $\pm$ 0.38 & 83.78 $\pm$ 0.26 & 13.19 $\pm$ 0.4 & 0.42 $\pm$ 0.41 \\
$LLAMA+Obj_{1,2}$ & \textbf{27.97 $\pm$ 0.11} & 84.47 $\pm$ 0.29 & 13.33 $\pm$ 0.2 & 0.43 $\pm$ 0.64 \\
$LLAMA+Obj_{1,3}$ & 27.14 $\pm$ 0.07 & 90.02 $\pm$ 0.4 & \textbf{13.11 $\pm$ 0.1} & 0.4 $\pm$ 0.35 \\
$LLAMA+Obj_{1,2,3}$ & 27.84 $\pm$ 0.22 & \textbf{91.07 $\pm$ 0.15} & \textbf{13.11 $\pm$ 0.4} & \textbf{0.43 $\pm$ 0.42} \\
\midrule
$T5$ & 27.44 $\pm$ 0.31 & 79.61 $\pm$ 0.33 & \textbf{1.43 $\pm$ 0.3} & 0.38 $\pm$ 0.46 \\
$T5+Obj_{1}$ & 27.61 $\pm$ 0.1 & 79.07 $\pm$ 0.28 & \textbf{1.43 $\pm$ 0.2} & 0.41 $\pm$ 0.35 \\
$T5+Obj_{1,2}$ & 28.19 $\pm$ 0.18 & 81.39 $\pm$ 0.46 & 1.44 $\pm$ 0.2 & 0.42 $\pm$ 0.51 \\
$T5+Obj_{1,3}$ & 27.52 $\pm$ 0.39 & 81.98 $\pm$ 0.37 & 1.44 $\pm$ 0.3 & 0.42 $\pm$ 0.24 \\
$T5+Obj_{1,2,3}$ & \textbf{28.25 $\pm$ 0.12} & \textbf{82.9 $\pm$ 0.27} & \textbf{1.43 $\pm$ 0.2} & \textbf{0.44 $\pm$ 0.36} \\

    \bottomrule
    \end{tabular}
    \caption{Quantitative evaluation of our decoding strategy for controlling toxicity in text generation (mean and standard deviation; three runs).
    }\label{tab:tab_1}
\end{table*}

\paragraph{Implementation details} We fine-tune the language models on an NVIDIA GeForce RTX GPU with 24GB of GPU RAM during 8 epochs. We set the learning rate to 0.0001 and the batch size to 4 for T5 and BART and to 1 for LLAMA. We use nucleus sampling \cite{DBLP:conf/iclr/HoltzmanBDFC20} with $p=0.9$ during inference. Compared with the commonly used beam search, nucleus sampling is more effective and can better prevent text degeneration \cite{DBLP:conf/iclr/HoltzmanBDFC20}. The matching between the generated interpretations and the human interpretations is done using the Hungarian algorithm. Our code is available here: \url{https://github.com/mtrusca/ToxicityControl}.

\begin{table*}[t]
    \centering
    \footnotesize
    \begin{tabular}{l cccc}
    \toprule
       Method & \multicolumn{1}{c}{$\text{$METEOR$}(\uparrow$)}  &  \multicolumn{1}{c}{$\text{$COMET$}(\uparrow$) } & \multicolumn{1}{c}{$\text{$Perplexity$}(\downarrow$)}  &  \multicolumn{1}{c}{$\text{$Correlation$}(\uparrow$) }  \\
      \midrule
$LLAMA+Obj_{1,3} (\lambda=.25)$ & 27.44 $\pm$ 0.09 & 88.93 $\pm$ 0.38 & 13.11 $\pm$ 0.1 & 0.41 $\pm$ 0.36 \\
$LLAMA+Obj_{1,3} (\lambda=.50)$ & 27.54 $\pm$ 0.28 & 89.93 $\pm$ 0.22 & 13.12 $\pm$ 0.2 & 0.4 $\pm$ 0.28 \\
$LLAMA+Obj_{1,3} (\lambda=.75)$ & 27.36 $\pm$ 0.38 & 90.44 $\pm$ 0.22 & 13.11 $\pm$ 0.1 & 0.41 $\pm$ 0.21 \\
$LLAMA+Obj_{1,3} (\lambda=1)$ & 27.14 $\pm$ 0.07 & 90.02 $\pm$ 0.4 & 13.12 $\pm$ 0.1 & 0.4 $\pm$ 0.35 \\
$LLAMA+Obj_{1,2, 3}$ & \textbf{27.84 $\pm$ 0.22} & \textbf{91.07 $\pm$ 0.15} & \textbf{13.11 $\pm$ 0.4} & \textbf{0.43 $\pm$ 0.42} \\

    \bottomrule
    \end{tabular}
    \caption{The effect of $\lambda$ on the decoding mechanism for toxicity control. While the first four models use a fixed $\lambda$, model $LLAMA+Obj_{1,2, 3}$ implements a decreasing $\lambda$ as the toxicity of the input sentence increases.
    }\label{tab:tab_2}
\end{table*}

\begin{table*}[h]
    \centering
    \footnotesize
    \begin{tabular}{l lc}
    \toprule
\textit{\textbf{Sentence}} & \textit{Over the past few years across Western nations, there have been several attempts by} & Tox.: 0.04 \\
& \textit{`Progressives' to gain power.} & \\
\midrule
\textbf{Ground-Truth} &  1. Politics that have stagnated are due to be contested. & Tox.: 0.03 \\
\textbf{Interpretations} & 2. Over the past few years across Western nations, there have been several endeavors by& Tox.: 0.04 \\
&`Progressives' to gain puissance. & \\
\midrule
\textbf{Generated} & 1. Western countries are not better than the Russians when it comes to politics. & Tox.: 0.03 \\
\textbf{Interpretations} & 2. In recent years across Western countries, many left-leaning politicians have been trying & Tox.: 0.08 \\
&to gain power & \\
\bottomrule
\toprule
\textit{\textbf{Sentence}} & \textit{Zoomers are inexperienced, idealistic do-gooders who are woke, and Boomers are world}& Tox: 0.47\\
&\textit{destroyers who consume everything without thinking, and will be the death of us all.} & \\
\midrule
\textbf{Ground-Truth} & 1. The young generation is too idealistic and not realistic while the older generation& Tox: 0.18 \\
\textbf{Interpretations}&consumes resources mindlessly. & \\
 & 2. Zoomers are callow, idealistic do-gooders who are woke, and Boomers are world& Tox: 0.45 \\
&destroyers who consume everything without mentally conceiving, and will be the& \\
&demise of us all. & \\
\midrule
\textbf{Generated} & 1. Millennials (zoomers) are naive idealists who are attuned to the environment and& Tox: 0.53 \\
\textbf{Interpretations} &boomers are extinctionists who consume everything without thought, and will ultimately& \\
&kill us. & \\
& 2. Zoomers and Boomers' lifestyles are completely different. & Tox: 0.19 \\
\midrule
\textbf{Generated} & 1. Zoomers and Boomers have different ideals on how to deal with the world. & Tox: 0.17 \\
\textbf{Interpretations} & 2. The writer seems to be pointing to a kind of inter-generational difference that motivates& Tox: 0.21 \\
\boldmath{($tox(s) = 0.2$)}&and polarizes extreme political movements. & \\

    \bottomrule
    \end{tabular}
    \caption{Examples from the OrigamIM test dataset that present toxicity behavior in $LLAMA+Obj_{1,2,3}$.
    }\label{tab:tab_3}
\end{table*}

\paragraph{Metrics} We use METEOR \cite{DBLP:conf/acl/BanerjeeL05} to measure the syntactic similarity between the human interpretations and the generated ones. We measure semantic similarity using COMET \cite{DBLP:conf/emnlp/ReiSFL20}. 
COMET is suitable for interpretation modeling because it was trained to recognize human preferences between correct and incorrect translations, which can be applied to the "translations" of meaning in interpretations. Additionally, COMET considers both the similarity between the generated interpretation and the human interpretation, as well as between the generated interpretation and the input sentence.
The third metric we report is perplexity, which shows the level of uncertainty the models have in predicting the generated interpretations. 
The final metric is the Spearman correlation computed between the toxicity scores of the generated interpretations and the scores of the human interpretations. 

\section{Results}

\paragraph{Quantitative analysis} Table \ref{tab:tab_1} presents the quantitative results of integrating our method into the text decoding of T5, LLAMA, and BART models. Syntactically, we notice that controlling toxicity in text generation consistently enhances the capacity of the models to generate interpretations similar to the input sentence. Analyzing METEOR scores, we observe that the implementation of the first objective has the strongest capacity to increase syntactic similarity, while the implementation of the other two objectives further enhances this similarity, as observed in the cases of LLAMA and T5. Regarding semantic similarity, the meaning of the input sentence is better preserved when toxicity is directly adjusted during decoding. When toxicity is controlled using all three objectives, COMET scores show a substantial increase compared to the results of the base models, with improvements of 4.10\% for BART, 5.54\% for LLAMA, and 4.04\% for T5.

Regarding perplexity, implementing our decoding strategy generally results in lower model uncertainty when generating the interpretations. Correlation scores further confirm that
the toxicity-controlled interpretations better capture the toxicity behavior observed in human interpretations than when toxicity is not controlled. Lastly, the results show overall improvement in the interpretation generation performance when all three objectives are enforced. 

To demonstrate that a variable $\lambda$ value (as required by \textbf{Objective 2}) is more advantageous than a fixed value, we evaluate our decoding strategy using different fixed $\lambda$ values ($\lambda = 0.25, 0.50, 0.75, 1$). As shown in Table \ref{tab:tab_2}, a variable $\lambda$ results in better manipulation of the toxicity level in the generated text and achieves higher semantic and syntactic similarity to the human interpretations, compared to when $\lambda$ is fixed.

\paragraph{Qualitative analysis} Table \ref{tab:tab_3} presents several interpretations generated by LLAMA using our decoding strategy. When the toxicity score of the input sentence is low, the generated interpretations are also non-toxic. However, this does not prevent LLAMA from being creative and discussing Russian politics in the context of Western political systems. Conversely, when the input sentences have a high level of toxicity, the generated interpretations either reflect the toxicity or produce milder interpretations. Note that we can moderate the toxicity of an input sentence by replacing its toxicity score $tox(s)$ with a lower value that allows generation of non-toxic interpretations (last line in Table~\ref{tab:tab_3}).

\section{Conclusion}
In this work, we proposed a modular decoding algorithm with three objectives designed to explicitly guide the generation of interpretations of out-of-context sentences. 
We showed that specifically constraining text decoding on toxicity brings generated interpretations closer to those written by humans. 
However, human interpretation is driven by many factors beyond toxicity like cultural background and personal experiences. We therefore strongly encourage future research to also consider these contextual factors when modeling the diverse ways in which a sentence's meaning is perceived.

\section*{Limitations}

Due to the external classifier used to detect toxicity, the ability to control the toxicity of our decoding strategy is strongly correlated with the data used to train the classifier. As a result, our strategy depends on the quality of the classifier's training data.


\section*{Ethical Considerations}
Our decoding method intentionally amplifies toxicity in certain generated interpretations to better replicate human interpretations of out-of-context sentences with varying levels of toxicity. While promoting toxicity in text generation may seem controversial, it is not inherently negative in all contexts. Minimizing or even entirely removing toxicity is crucial for applications like customer service, education, or mental health support -- where safety and ethics are non-negotiable. However, some systems actually benefit from the ability to produce texts with varying degrees of toxicity. For example, explicitly highlighting toxicity in generated text can help improve content filtering systems and facilitate better detection of harmful language. As such, we believe that developing methods for the controlled and adaptable regulation of toxic language is valuable. Nevertheless, it is important to exercise caution in designing and implementing these methods to ensure they are used responsibly and ethically.

\section*{Acknowledgements}
This work has been funded by the Research Foundation - Flanders (FWO) under grant G0L0822N through the CHIST-ERA iTRUST project.

\bibliography{anthology,custom}
\bibliographystyle{acl_natbib}

\appendix




\end{document}